\documentclass{article}
\usepackage{ijcai25}

\usepackage{fontawesome5}
\usepackage{enumitem}
\usepackage{times}
\usepackage{soul}
\usepackage{url}
\usepackage[hidelinks]{hyperref}
\usepackage[utf8]{inputenc}
\usepackage[small]{caption}
\usepackage{graphicx}
\usepackage{amsmath}
\usepackage{natbib}
\usepackage{amsthm}
\usepackage{booktabs}
\usepackage{algorithm}
\usepackage{algorithmic}
\usepackage[switch]{lineno}
\usepackage{txfonts}

\title{An AI-Based Public Health Data Monitoring System}

\author{
Ananya Joshi$^1$
\and
Nolan Gormley$^1$\and
Richa Gadgil$^1$\and
Tina Townes$^1$\and
Roni Rosenfeld$^1$\and
Bryan Wilder$^1$
\affiliations
$^1$Carnegie Mellon University\\
\emails
\{aajoshi, ngormley\}@andrew.cmu.edu
}

\begin{document}

\maketitle

\begin{abstract}
Public health experts need scalable approaches to monitor large volumes of health data (e.g., cases, hospitalizations, deaths) for outbreaks or data quality issues. Traditional \textit{alert-based} monitoring systems struggle with modern public health data  monitoring systems for several reasons, including that alerting thresholds need to be constantly reset and the data volumes may cause application lag. Instead, we propose a \textit{ranking-based} monitoring paradigm that leverages new AI anomaly detection methods. Through a multi-year interdisciplinary collaboration, the resulting system has been deployed at a national organization to monitor up to 5,000,000 data points daily. A three-month longitudinal deployed evaluation revealed a significant improvement in monitoring objectives, with a 54x increase in reviewer speed efficiency compared to traditional alert-based methods. This work highlights the potential of human-centered AI to transform public health decision-making.
\end{abstract}

\maketitle
\section{Introduction}

\begin{figure}
\centering
\includegraphics[width=0.9\linewidth]{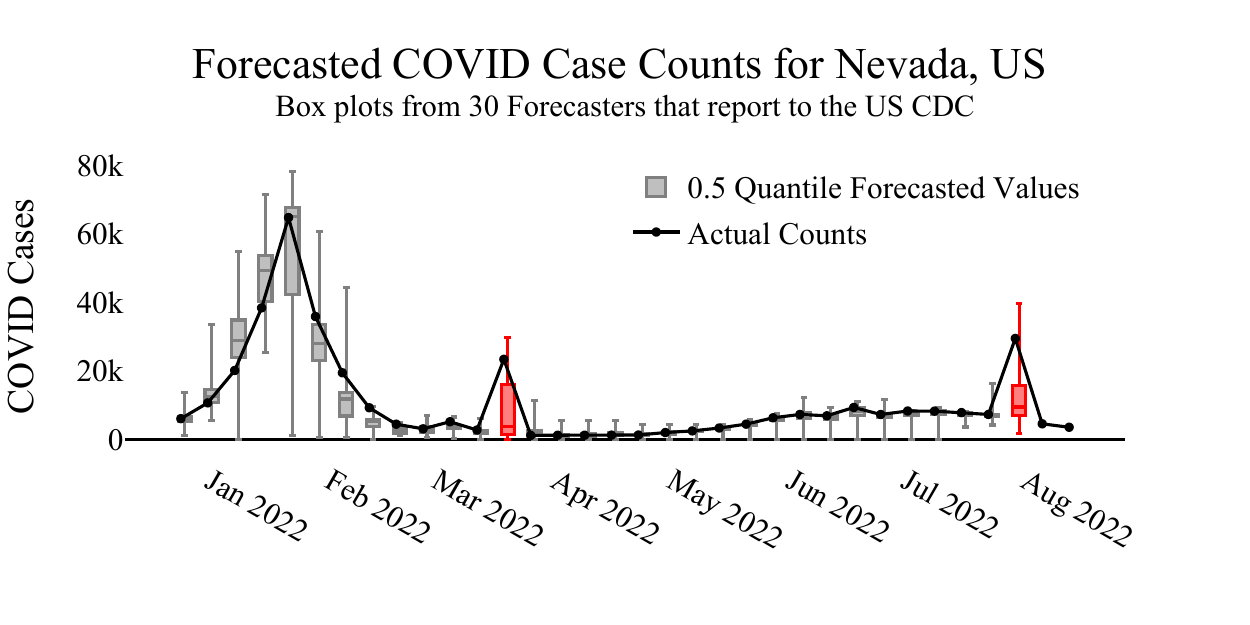}
\caption{Data monitoring catches the data quality changes in case counts, shown by the large spikes when cases were trending down, which resulted in similar spikes for predicted counts (red) from multiple forecasts.
 }
\label{fig:monitoring}
\end{figure}

 Public health data monitoring systems are critical for detecting outbreaks and ensuring data quality for downstream applications (Fig \ref{fig:monitoring}), but traditional alert-based systems struggle to detect these outlying data events with modern, high-volume, heterogeneous data streams \cite{shmueli2010statistical}.

To improve timely and informative public health actions, data modernization efforts in the past decade have led to an exponential increase in data volume, complexity, and heterogeneity \citep{whostrat}, especially at national-level data curation organizations like \textbf{Delphi}. Ironically, these investments in data modernization have been \textit{incompatible} with existing public health monitoring systems, reducing the utility of this investment\footnote{Ethical Approval / Approving Organization: This study was IRB approved.}. Specifically, most organizations used alerting-based monitoring systems -- where the computational data processing methods often involved defining specific types of alerts for data patterns that matched data quality issues or outbreaks \cite{ dong2022johns}. However, these definitions could not adapt to the wide variety of public health data sources, fluctuating quality, and increasing volume in the modern data —- unlike the relatively small, static data pipelines of the past. After observing the failure modes using the alerting paradigm, as we describe in this paper, we, a group of researchers at Delphi, designed a novel monitoring system that addresses the core limitations of the prior system via a newly available AI-based ranking paradigm. \\\\

\noindent{\textbf{Goal: } The `gold standard' of a successful public health data monitoring system is when data reviewers 1) can quickly find as many high-priority events and 2) increase their effeciency.} \\\\
\noindent{\textbf{Limitations of Current Alerting Paradigm}}
Existing alerting-based monitoring systems, where data that meets a statistical definition produces an alert, are incompatible with modern data because they: 
\begin{enumerate}
    \item {Are stagnant to data changes (Computational) :} At scale, the alerting algorithm parameters needed to be manually adjusted across millions of data streams because of the fluctuating public health volume and measurement standards. This requirement for manual tuning negates the scalability benefits of automation.
    \item {Constrain situational awareness (Engineering)}: At scale, individual data points need to be inspected in the \textit{context} of the larger public health data available \citep{burkom2017evolution}. However, the volume of alerts and constraints on relevant context (e.g. alerts from cases and deaths cannot be combined because cases usually lead deaths so the alerts may be unrelated) increase engineering complexity and introduce application latency.
    \item {\textbf{Disincentivize engagement} (Reviwer Actions):} Most importantly, reviewers using these applications often had to sort through an overwhelming and unusable number of data alerts (e.g. 35,000+ alerts \citep{coletta2019can}).  Without the context they need to engage with the system from 2., they cannot meet their minimum goal of correctly analyzing high priority events. 
\end{enumerate}

In response to these challenges, we, a research team at Delphi, designed a ranking-based AI system that leverages the strengths and constraints of multiple public health stakeholders, that has been used for the past two years. In the process, we conducted a multi-year collaboration involving public health data reviewers, engineers, and computer scientists, identified aspects of the process that were well suited for an AI approach, and developed an approach to evaluate it in practice. In the following sections, we describe the process of developing this system via: \\
$\bullet$ {\textbf{Sec 4:}} Modernizing Data Monitoring with AI: opportunities and constraints on automation. \\
$\bullet$  {\textbf{Sec 5:}} Goal-oriented system design guidelines  \\
$\bullet$ {\textbf{Sec 6,7:}      Measurement, evaluation, and deployment strategies for empirical validation

\begin{figure}
\includegraphics[width=\linewidth]{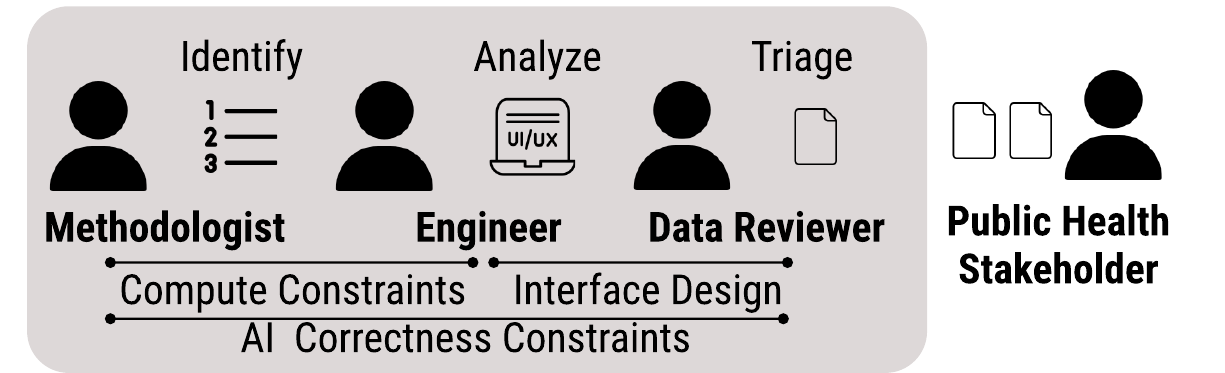}
\caption{Data reviewers triage notable events from public health data for stakeholders. Our system is powered by an AI-ranking method that identifies relevant events and formalizes how different stakeholders interact directly or indirectly with the AI method. 
 }
 \label{fig:preteaser}
  \end{figure}

The final system (Fig. \ref{fig:preteaser}) centers improved human interactions through an unsupervised, ranking-based, AI paradigm. To test the impact of this approach on the system over time, we documented the performance of this human-AI collaboration over 3 months, where it was subject to the fluctuations and changes typical in public health data. During this time, we also added interaction modalities for stakeholders that interact with the AI outputs to test and support the data monitoring system.  We \textit{preregistered} the Github commits and evaluation strategy on OSF before experiments began. Our results show that the ranking paradigm outperformed the alerting paradigm on core qualitative and quantitative metrics designed by reviewers that demonstrate the improved 1) effectiveness of the ranking paradigm in practice, 2) efficiency of reviewers using the system, and 3) overall experience with the system, which support data monitoring in the modern setting.

\section{Related Work}
The challenges encountered in public health data monitoring closely resemble those in other critical domains requiring large-scale data monitoring (e.g., environmental, financial, and agricultural sectors), especially because these fields depend on domain expertise for reasons such as liability and contextual interpretation. Numerous systems exist to help these types of  experts understand relationships within large volumes of temporal data \citep{kesavan2020visual}. However, most of these systems focus on surfacing aggregate trends rather than monitoring individual data points in the \textit{context} of these trends. As modern data streams become more complex, these types of general-purpose data analysis systems become inadequate for the following reasons: 

\textbf{{Stagnant to Data Changes:} }
Alerting systems at scale suffer from well-documented issues, including statistical inaccuracies \citep{joshi2024outlier, shmueli2010statistical} and overwhelming alert volumes \citep{hurt2005syndromic}. Even the most sophisticated adaptive mechanisms \citep{burkom2017evolution} require infeasible, constant retuning to provide a reasonable set of high quality alerts due to the fluctuating quality of data.

\textbf{{Missing Situational Awareness:} }
Some systems fuse data across visual interfaces \citep{deng2023survey, cao2017voila, vajiac2022trafficvis}. While generally effective in typical HCI and visualization settings, these approaches are incompatible with public health data due to its shifting lead times across indicators and complex correlation structures \citep{Lakdawala_2023}. Static correlation-based techniques often fail in this context \citep{fan2014challenges, hilda2016review}.  

Additionally, the nonstationary and hierarchical nature of public health data \citep{reinhart2021open} introduces constrained relationships that must be accounted for to prevent spurious correlations. This failure mode is very likely in public health given that data quality varies across univariate data streams, within indicators across regions, and between different indicators for the same region. This variability also complicates \textbf{directly} comparing across data dimensions, impacting which limited subset of events are detected by these methods. Consequently, many existing methods that are highly parameter-sensitive, computationally expensive \citep{teng2017anomaly, kesavan2020visual}, or require high visual literacy cannot work in these setting. These limitations are prominent even in more recent ranking-based approaches \citep{ding2023explanation, montambault2022pixal}, which often treat all data dimensions similarly—an inappropriate assumption for public health data, where temporal and geospatial dimensions are far more important than other dimensions.

\textbf{{Disincentivizes Engagement:}} 
Systems that generate overwhelming numbers of alerts can lead to information overload, preventing thorough review. Exploratory filtering across dimensions and indicators (e.g., cases, deaths, hospitalizations) \citep{preim2020survey, chen2010data, maciejewski2009visual} requires users to navigate millions of filter combinations to inspect relevant data points. While batching alerts \citep{essencepremier}\footnote{For example, public health alerts can be aggregated over data provider, indicator, geography, time, revision history, or population.} can reduce alert volume, this approach risks improper aggregation. For example, alerts for cases and deaths cannot be combined due to their distinct lag structures and event correlations. Further, contemporary drill-down techniques demand extensive manual inspection \citep{hurt2005syndromic}, further slowing the review process.  

These challenges are exacerbated in modern public health monitoring. First, when combined with high-volume, low-quality data—characterized by nonstationarity, noise, missing values, and inconsistencies over long timespans -- the aforementioned theoretical limitations of alerting methods are realized.  Second, as existing systems using the alerting paradigm were \textit{designed} to only be used on a local scale \citep{chen2010infectious}, anecdotally where \textbf{all} data could still be analyzed manually, they were never built to support situational awareness. As a result, the need for improved monitoring systems has been widely recognized by public health practitioners \citep{hopkins2017practitioner}, with limitations persisting across both historical \citep{chen2010infectious} and modern public health monitoring systems.  

A final challenge in designing monitoring systems is that academic research on disease monitoring systems has predominantly focused on improving  methods for data identification in isolation \citep{buckeridge2007outbreak}, including recent AI approaches. Unfortunately, this has not always translated into practical adoption \citep{shmueli2010statistical}. Instead, we adopt an interdisciplinary systems design approach, inspired by:
    ``The current disconnect among algorithm developers, implementers, and users has ... foster[ed] distrust in statistical monitoring and in biosurveillance itself" \citep{shmueli2010statistical}.




\section{Monitoring and Study Background} 
We identified the interests of the following involved parties through interviews and discussions over 2 years. These following insights were synthesized from interviews and interactions with 5 public health departments.

A core responsibility of the Delphi Research Group at Carnegie Mellon University is to its public health stakeholders. Since 2020, these stakeholders, including public health decision makers, modelers, and journalists wanted to find the `needle in the data haystack' -- or important public health events from Delphi's data relevant to \textit{their} regions. They were concerned that a black-box AI approach might overlook their regions. This was particularly important since smaller populations have historically been deprioritized because the underlying alerting methods penalized small counts, leading to systematic underreporting.

Additionally, a fully closed-looop AI based approach for monitoring was unacceptable. Human reviewers for public health data provide critical context for missing and rapidly changing data in public health settings, especially in cases where the reporting structure fundamentally changes in a way that would take several days for an AI model to automatically adapt. For instance, a sharp rise in hospitalizations may appear to signal an outbreak to an AI method, but a human could contextualize this given other changes, like a new policy about how hospitalizations are recorded. Given that liability was also a concern, the AI method could not operate independently and all outputs needed to be `human-verified'. The following parties (number involved) were then tasked with meeting stakeholder requirements for monitoring.

\begin{itemize}[label=\faUser]
     \item \textbf{Data Reviewers (3):} They perform the daily data monitoring. To meet their aformentioned output goal from the monitoring system their key performance indicators (KPIs) around efficiency (how quickly they identify events) and efficacy (accuracy of events identified). They need interfaces that are responsive, provide necessary context, and are easy to learn with minimal onboarding \citep{ooge2022explaining}.

    \item \textbf{Engineers (3):} Building a functional monitoring system is considerable effort, especially one that serves and visualizes large volumes of data quickly. The three engineers on this project needed to find a balance between data processing and system responsiveness so that data reviewers would not lose focus or attention\footnote{The past 200 days of data are queried from a virtual database in MySQL v.8. The event ranking method, written in Python3.10, is run separately on a 3.3GHz Intel Core i7 machine with Pop!OS daily with a cron job; the results are stored in a different MySQL v.8 database.}.  
    \item \textbf{Computer Scientist (3):} Few monitoring methods are appropriate for public health data because it is noisy, nonstationary, and has dynamic correlation structures. Existing methods generally have core statistical problems, which should be avoided in new design. 
\end{itemize}

After this initial information-gathering phase, a core team of 2 data reviewers, 2 engineers, 2 computer scientists was tasked with deployment. This core group collaborated for 18 months, meeting at least biweekly to refine the \textbf{deployed} system and interface.

\section{Modernizing Data Monitoring with AI}

\begin{figure}
  \centering
\includegraphics[width=0.8\columnwidth]{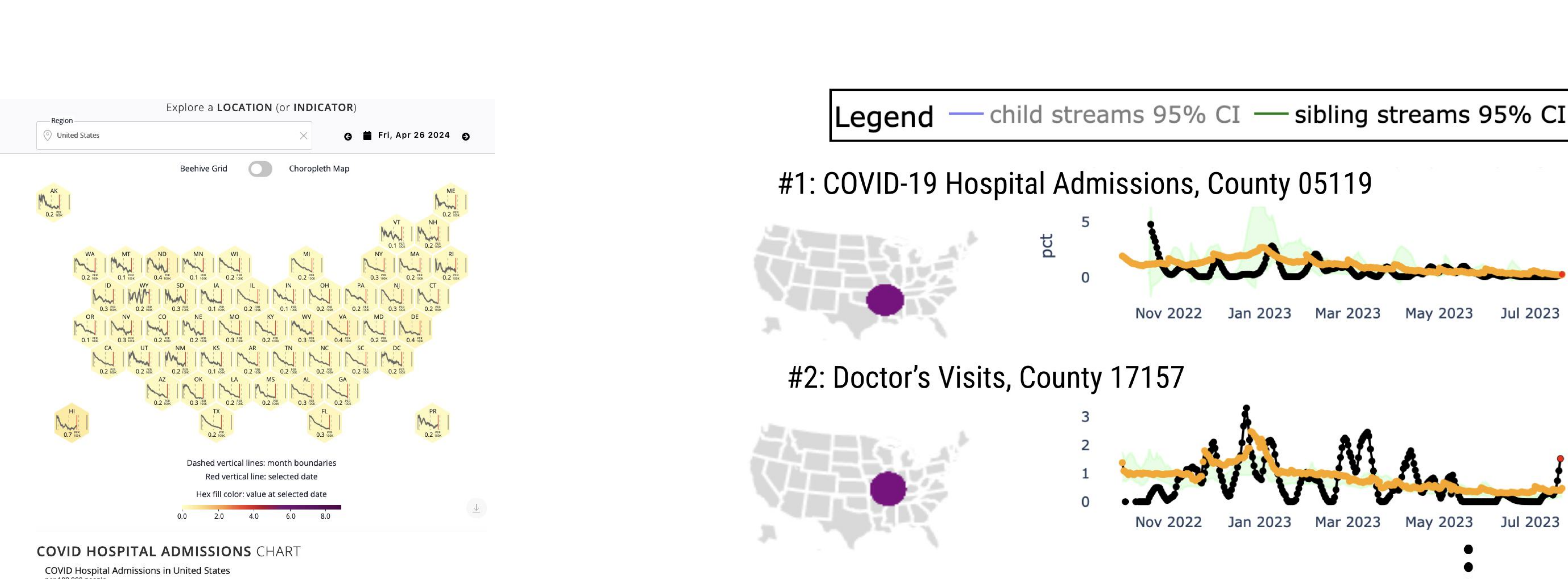} 
\caption{\textbf{A.} Baseline 1 (\textbf{deployed}) used to document events for stakeholder manually, but subject to the drill-down fallacy across dimensional layers. \textbf{B.} Baseline 2 showed a limited set of ranked data streams but was not used to document events.}
\label{fig:oldstreams}
\end{figure}

Based on our initial observational interviews, public health data reviewers perform three actions when monitoring data (Fig. \ref{fig:preteaser}):  
\begin{itemize}
    \item \textit{Identify unexpected (anomalous) data points.} These may indicate either reporting errors or early signs of an outbreak. Traditionally, this has been done manually or by using an alerting method.  
    \item \textit{Contextualize anomalies.} Reviewers assess anomalies within a broader situational context to ensure meaningful events are not overlooked. This triaging process helps them prioritize data for stakeholders.  
    \item \textit{Decide whether to continue reviewing.} This decision balances the reviewing urgency with the reviewer's attention capacity and capability to perform high-quality triaging. This is often based on the severity of recent triaged data points. 
\end{itemize}

As described, alerting systems fail to support these tasks effectively at scale. 
\subsection{Alerting-Paradigm Approach (Baseline 1)}
It was not immediately clear that an AI approach was appropriate or needed given that public health stakeholders wanted contextualized, expert-verified data. Our tested alerting based approaches included:

{1.} An approach similar to DHIS2 where we set alerts for z-scores calculated using an adaptive rolling window. Unfortunately, we ran into the same problem as \citep{coletta2019can} where there were tens of thousands of alerts daily that, when reviewed, provided little reason to remain engaged with the system (Statistical Limitation).

{2.} Batching alerts across different contextually meaningful dimensions (usually by indicator \citep{sarikaya2018we, essencepremier}) to support situational awareness masked important data \citep{pena2016fast} because it was difficult to know in advance which dimension these alerts should be batched along to be the most meaningful (Reviewer Limitation). 

3. \textbf{Baseline 1: \textit{Deployed Alerting System (120 weeks)}} Without a reliable automated system, reviewers reverted to {manual data inspection} using exploratory tools (Fig. \ref{fig:oldstreams}A). However, identifying anomalies through visual inspection was inefficient and error-prone \citep{joshi2023computationally, sibolla2018framework, hurt2005syndromic}. Further, aggregation strategies often led to a \textit{drill-down fallacy} \citep{lee2019avoiding}, where reviewers could still end up needing to explore a combinatorial number of dimensions to locate the data event that corresponded to the visual alert. 

\subsection{Alerting-Based Monitoring Fallacies}

Reviewer experiences with Baseline 1 became increasingly negative and echoed those of other public health professionals \citep{coletta2019can}. As expected by the public health stakeholders, reviewers missed important events, especially those in smaller regions and outside of the indicators that are displayed first on the interface. Then, between a) randomly choosing data filters, b) using multiple clicks to drill down to the raw data level, c) finding the appropriate regional tier responsible for the event by trial and error, and d) scrolling to compare indicator behavior across signals, whether relevant or not, reviewers became fatigued. 

Baseline 1's failure was the outcome of an alerting paradigm. When reviewers were presented with thousands of alerts to review daily, reviewers abandoned the statistical alerting approach and instead \textit{preferred the manual inspection method}, where they would review data based on their intuition (e.g. what they heard on the news). This was in \textit{conflict} with some stakeholders expectations that all regions have the opportunity to be inspected in a systematic fashion. 

Additionally, the computational scientists could not identify any alerting-based methods that could adapt to the nonstationarity and fluctuation volume of the data consistently enough without requiring constant parameter tuning, let alone one lightweight enough to meet the engineering constraints. After several initial design iterations, it became clear that there were two limitations to alerting: 

\textbf{Perscriptive Alerting Fallacy:} Many monitoring methods require parameter tuning over time to reflect changes in data dynamics. This tuning is in an effort to align the alerts with user expectations, but in doing so, often only catches the same types of anomalous data points a reviewer already knows about. In effect by focusing on identifying the known classes of unexpected data points, there are many other types that go undetected. 

Instead of deploying methods that require human approaches which are strongly prescriptive for data event detection, these systems \textit{should} be framed as measuring deviance from expected data. This is especially important because data reviewers cannot not determine if the alerts they were inspecting were not interesting because it was a calm day \textbf{or} because the alerting thresholds were tuned incorrectly and needed to be fixed, limiting their situational awareness and trust in the system.

\textbf{Threshold Fallacy:} The nature of the alerting paradigm is that while data points in a single stream can be prioritized, there is no standard way to do so across millions of data steams. Thus, the range of high priority and low priority ``alarms" are treated the same -- even if there is a difference, which there usually is from tens of thousands of alarms. Reviewers, already with limited time and energy need to decide if continuing to inspect data is worth it. Under the alerting paradigm, reviewers could not determine whether they were seeing a low-alert day (i.e., truly calm public health conditions) or if the alerting system was miscalibrated—further diminishing engagement.  

\subsection{Shift to AI-Based Ranking}

To address these issues, we designed a system around \textbf{anomaly ranking} rather than threshold-based anomaly alerting. Unlike binary alerts, a ranking-based AI system:  
\begin{itemize}
    \item Scores data points based on deviation from expectation (addressing prescriptive alerting).  
    \item Helps reviewers decide whether further inspection is warranted (addressing threshold).  
    \item Aligns with how human experts naturally assess data (meets observations).  
\end{itemize}

This was especially important as reviewers were not mandated to use the system for a specific amount of time or in any forced way. Instead, their investment in monitoring was directly tied to how important they thought the day's alerts were. Of the few \textit{anomaly ranking methods} that support this setting, the most appporpriate was a multiple-univariate outlier detection approach for monitoring \citep{blazquez2021review}, specifically, a validated human-in-the-loop, linear-time monitoring method \citep{joshi2024outlier}. This method relies on the user to define \textit{data expectations} and scores data points at scale, making it an appropriate class of AI methods to base the monitoring system in.  

\textbf{Baseline 2: Method AI Approach} The selected method was run in isolation over 30 weeks with minimal changes to a threadbare system. While this toy approach was not used to document events for stakeholders, it serves as a second type of baseline for just the AI ranking approach in isolation vs. part of a human-in-the-loop system over time. In this approach, we displayed interactive line plots in a static HTML file for the top-k data streams (Fig. \ref{fig:oldstreams}B).

With the underlying AI method in place, the next step was to design a robust interaction model that met reviewers needs and could document events for stakeholders. 

\section{Design Guidelines and Expectations}

To design interaction modalities with the AI ranking method, the group developed the following design guidelines: 

\begin{enumerate}[label=(C\arabic*)]
\item \textbf{Meaningful Interaction with AI Method:} To avoid the circular logic of the prescriptive alerting fallacy, changes computer scientists make to the method had needed to be translated directly from interactions with the engineers or data reviewers.  This ensured that changes were grounded in real-world use cases rather than arbitrary threshold adjustments.  
\item \textbf{Providing Situational Awareness:} To address the issue of misleading alert inspection 
the system interface components must provide context to analyze and triage data. Here it is critical to balance avoiding interface and visualization oversimplification with cognitive overload to prevent data misinterpretation. 
\item \textbf{Engagement.} Engagement is critical to sustained system use, but can be easily diminished. One challenge previously observed was \textit{algorithmic fatigue}, where reviewers become less engaged with the system, e.g. due to the volume of alerts. 
\end{enumerate}

\section{Human-AI Interaction Modalities}

In consideration of the previously outlined constraints and design guidelines, the team developed the following modalities for human interaction with the AI method (Fig. \ref{fig:preteaser}). In general, the AI method is designed to incorporate correctness and computational constraints specified by the methodologist. These constraints or parameters are integrated into the method to balance computational efficiency with accuracy. The methodologist collaborates with both engineers and data reviewers to iteratively refine these constraints, ensuring that the system meets their needs. For example, changes in the data quality of public health require modifications to the AI method’s inputs (e.g. adjusting aggregation policies, incorporating multivariate forecasting approaches to account for signal lag or correlation, or interpolating missing data). Modifications to the method typically occur on a monthly basis, aligning with the frequency at which public health data collection processes and statistical properties evolve. 

Beyond these interaction modalities, reviewer analysis and triaging actions require an \textit{interface} that appropriately reflect the AI method’s outputs. Together, the AI method, interaction modalities, and the interface constitute the system design for our modern monitoring system. To evaluate the effectiveness of this system design beyond the AI method evaluation alone (Baseline 2), the group focused on strategies to enhance situational awareness and user engagement. Additionally, the team developed a meta-deployment and evaluation framework to assess Human-AI interaction effectiveness \textit{over time}. In this approach, we make sequential changes to the interface that reflect reviewer needs in ways that ensure reviewers have sufficient time to adapt to system changes and provide informed feedback \citep{carroll2014visualization, janes2013effective}.


\section{Interface $\&$ Interaction  Experiments}
\label{sec:review} The basic revised monitoring system takes the AI ranked list of data streams and (Fig. \ref{fig:pan}) segments the data so the temporal dimension is emphasized in analysis.  Reviewers can easily expand each data row, which includes a custom map to orient the reviewer and other stream properties. Each row also contains an interactive line plot with data streams. Notably, \textit{contextualization} for data across geographical tiers is controlled by legend items the reviewer can toggle to see  data streams in the same tier that share the same regional parent (i.e., a sibling stream), as well as parent streams and child streams with a 95 \% CI. To standardize the triaging process, they create a record corresponding to the type of event (e.g., a provider issue), its severity (low, medium, or high), and if the point identified was the source of the event (yes/no). 

Notably, towards situational awareness, reviewers also used an interface to record \textit{meta-events} that combine multiple events from individual data points into informative, higher-level phenomena. These meta-events are based on  hypotheses across events that reviewers want to investigate, and are very informative to Delphi's stakeholders if there are a large number of pressing events. There were 3 modifications to the basic system that were developed and evaluated (M1, M2, M3):

\textbf{M1: Data Point Filter}
Given the choice to focus on the temporal dimension of the data as the primary interaction segment, we need to design how users will access other data segments. The number of possible filtering combinations in this setting, especially considering the desire for multiple filters (including exclusions) per category of data provider, indicator, and geographic region. Given this, we use this step to validate if the performance simple filtering strategy is appropriate for data review. 

\textbf{M2: Situational Awareness Panels}
Our approach supports situational awareness via two displays of event scores segmented and aggregated across geography and indicator. This is only possible because event scores can be compared across these dimensions wheras raw values cannot due to spatial heterogeneity of public health data. Scores $\textbf{s}\text{(x)}$ across different spatial tiers $e \in \mathcal{E}$ where $\mathcal{E}$ includes county, state, and nation) are aggregated across indicators $i \in I$, and time (T:T-7). For the map, the choropleth color value $c$ for each county (on a scale of 0 to 1) is calculated using $\forall r \in \mathcal{R}, \quad c\text{(r)} = \frac{\sum_{e \in \mathcal{E}}\frac{\sum_{i \in I} \log \text{(}\bar{s}\text{(}x_{i,e\text{(r)},T-7:T}\text{)+}  1\text{)(}/log\text{(w)} }{|I|}}{|\mathcal{E}|}$

, where $e\text{(r)}$ is the region that subregion $r$ belongs to at tier $e$. The log scores make the more extreme events appear more clearly on the map. The indicator display scores are calculated using  $\frac{\sum_{r \in \mathcal{R}}\text{(}s\text{(}x_{i,r}\text{)}\text{(}T-7:T\text{))}}{|\mathcal{R}|}$.


\textbf{M3: Visualizing Data Evolution}
Finally, as data is revised over time, the historical event scores also change with the additional data availability and presence of new events. Capturing this evolution of event scores across historical revisions communicates the uncertainty \citep{carroll2014visualization} of the event scores over time to reviewers. Our approach for capturing the evolution event scores is inspired by the design of industrial stagger charts \citep{grove2015high}. It involves calculating the rolling mean and standard deviation over time, across data revisions, via Welford's online methods \citep{welford1962note}: $\bar{x}_{T} = \frac{\bar{x}_{T-1}\times(T-1) + x_T}{T}$; $\varv_{T} = \frac{\varv_{T-1} - (\bar{x}_{T-1} - x_T)\times(\bar{x}_{T}- x_T)}{T}$. We include a 1D heat map under the interactive time series plot to give reviewers the context of the event history and provide the average variance score across time. These help the reviewer understand the volatility in event scores over time, as shown in Figure \ref{fig:pan}.

\begin{figure}
    \centering
\includegraphics[width=0.8\columnwidth]{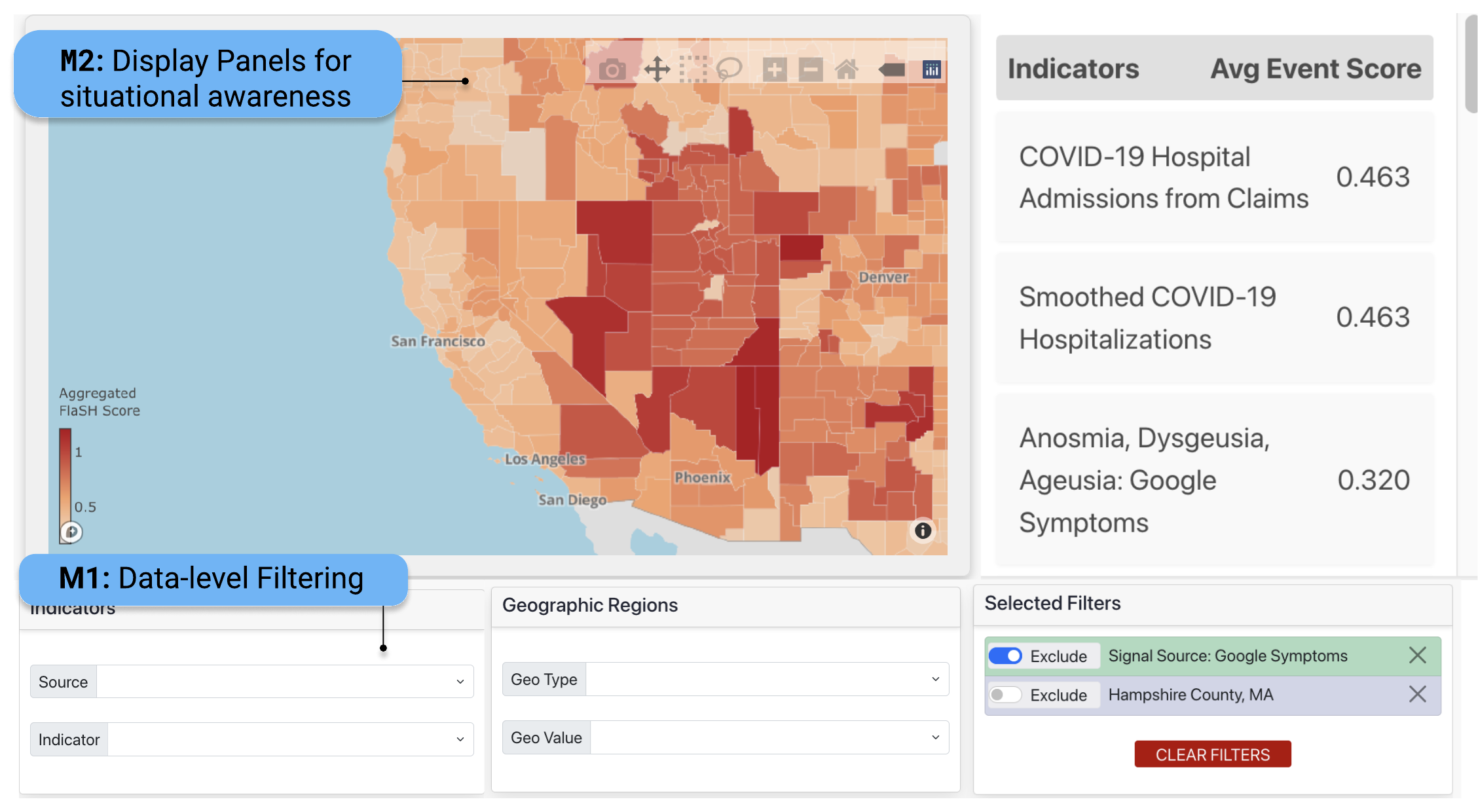}
     \caption{The initial displays support situational awareness and help reviewers get a sense of where data events may be found \textbf{(M2)}. This can help them guide their review via the data level filters (\textbf{M1)}. \textbf{M3} captures the impact of data evolution on event scores for each data row, and is shown on Fig \ref{fig:pan}.}
  \label{fig:init}
\end{figure} 
\begin{figure}
    \centering
\includegraphics[width=0.8\columnwidth]{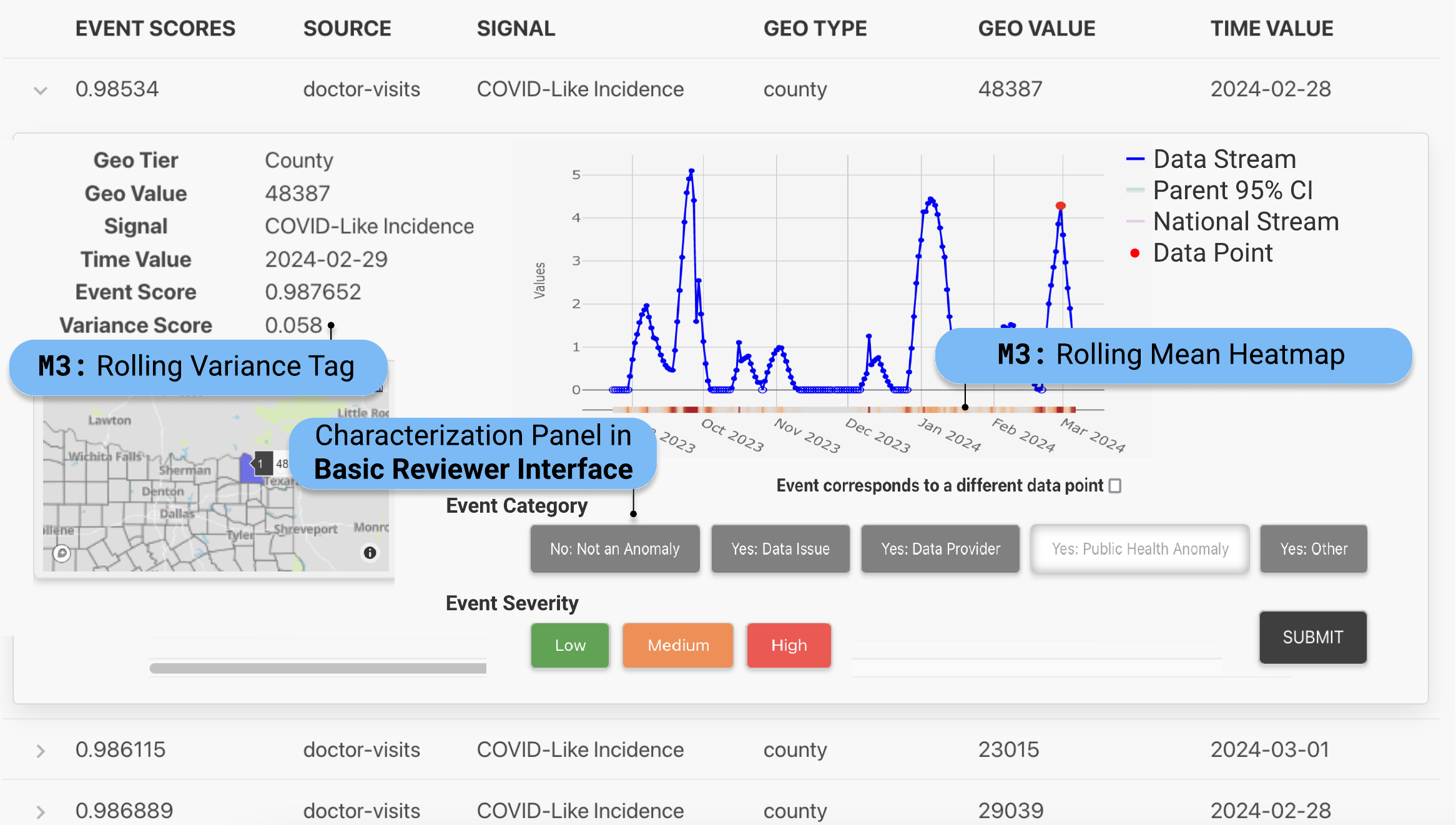}
 \caption{For mechanism \textbf{M3}, we added a tag with the rolling variance and a 1D heat-map of the rolling mean event scores so that reviewers have an intuition for how event severity changes over data revisions and time.}
  \label{fig:pan}
\end{figure}

\section{Evaluation}
Evaluating public health monitoring systems generally remains a challenge \citep{hyllestad2021effectiveness}. In particular,  longitudinal studies across large volumes of changing data remain under-discussed, despite their importance for adoption. As supported by Delphi's stakeholders, longitudinal studies are especially important for this data streaming evaluation because the daily reviewing load and the number of daily events vary. Typically, monitoring systems that initially perform well degrade over time, reducing reviewer trust and utility of these systems. Given this gap, our longitudinal, sequential evaluation uses metrics corresponding to key performance indicators (KPIs) data reviewers are evaluated against in meeting their intial goals, which we \textit{preregistered} on OSF before any evaluation began.

\noindent\textbf{Data Reviewer KPIs: }
Reflecting the initial `gold standard' goal of correctly analyzing high-priority events effeciently, we used the following types of metrics: 
\begin{itemize}
    \item \textit{Efficiency metrics:} (how quickly) time per row, number of events recorded per session
    \item \textit{Efficacy metrics:} (how well) number of events that were later revised, number of \textit{meta-events} recorded.
    \item \textit{Output metrics:} filter use, \% of times that the method identified data point was not the source of the event, and the distribution of the reviewer characterizations.
\end{itemize}

 To \textbf{contextualize} these numbers, we also include excerpts from a reflection that expert data reviewers published in a blog. Each evaluation phase took the standard public health timeline of 3-4 weeks \citep{biggerstaff2018results}.

\begin{figure}
 \centering
 \includegraphics[width=0.7\columnwidth]{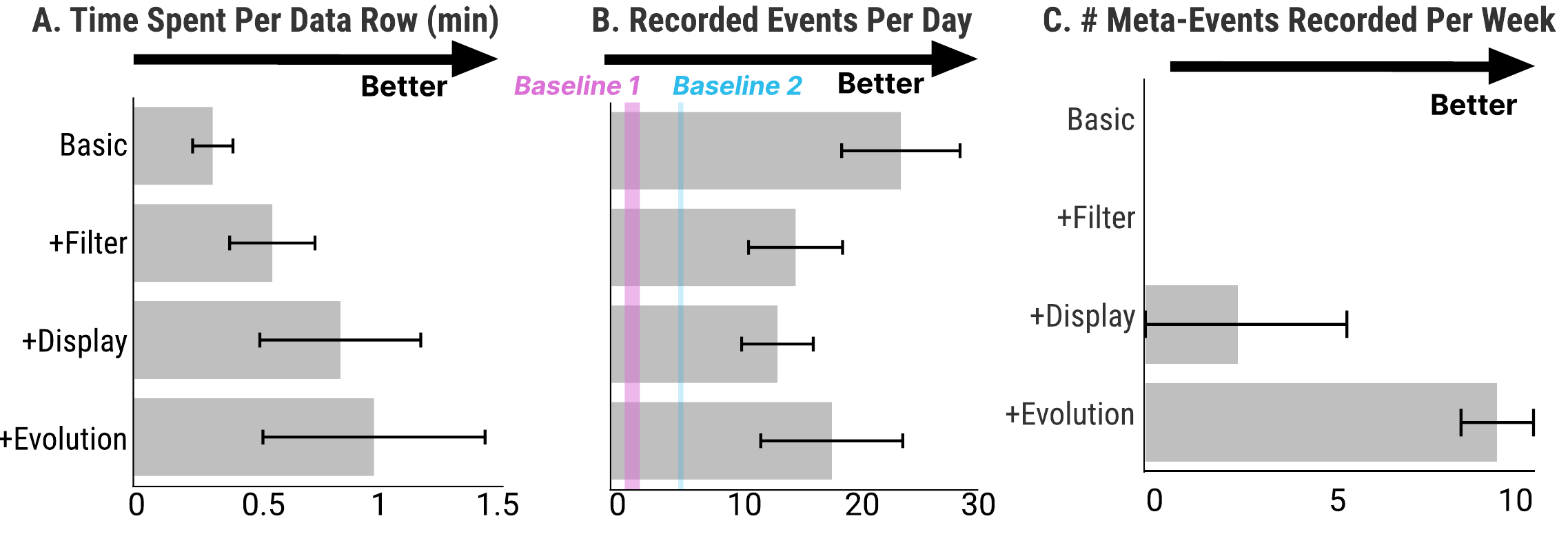}
 \caption{\textbf{A.} Reviewer engagement, displayed here with 95\% CI bars, increased with each added modification. Baselines had no comparable metrics. \textbf{B.} Reviewers  recorded significantly more events on average than the prior baselines. \textbf{C.} More meta-events were identified after situational awareness displays were added.}
 \label{fig:metrics}
\end{figure}

Efficiency metrics quantified a) how long reviewers interacted with each data row and b) the number of recorded events per day. As Fig. \ref{fig:metrics}A. shows, reviewers generally spent more time per row after each modification, particularly our choice of filters and adding display panels for situational awareness. This suggests that these modifications allowed reviewers to analyze the data deeply: \textit{``[the system] allow[s] me to devote more of my time and efforts to assessing points of interest."}

Reviewers also recorded far more events on average than with the prior baselines (Fig. \ref{fig:metrics}B.); reviewers were \textbf{54x} faster on average than while using the exploratory system in {Baseline 1}, and \textbf{5x} faster than the {Baseline 2} when recording events/minute. These metrics are contextualized by the reviewer:\textit{``[With the prior approaches], I was spending a good amount of time scrolling, manually sorting, documenting, and searching for specific [event] reports I wanted to examine rather than focusing solely on identifying, marking, and analyzing [events]."}

On efficacy, reviewers identify high quality events using the triaging system. If reviewers make a mistake and wish to correct a recorded event, they can easily update the record. In the past, this was frequently used as there are multiple informative external sources of outbreaks that reviewers contextualize against. While historically, this led to edits ({Baseline 2}'s responses had at least 3 edits across a similar experimentation timeline), there were no edits during the duration of this experiment. More importantly, reviewers identified meta-events when they could investigate patterns in the events that suggested higher-level phenomena. For example, a reviewer identified the following meta-event: 
\textit{``Several counties in Puerto Rico are repeatedly experiencing sudden upward trending, [respiratory illness] spikes, this month."}. No such meta-events were recorded for {Baseline 1} and only 2 were recorded for {Baseline 2}, as shown in Fig. \ref{fig:metrics}.

Finally, reviewers also seem to have a positive experience with this system, sharing: \textit{``the updated [triaging system] now enables me to [make meta-events] for exciting [events], trends and other issues of importance, and maintain these notes in an organized, searchable fashion within the platform."}
In a quality assurance check, these meta events were re-analyzed and corresponded with notable public health events.

\begin{figure}
 \centering
\includegraphics[width=\columnwidth]{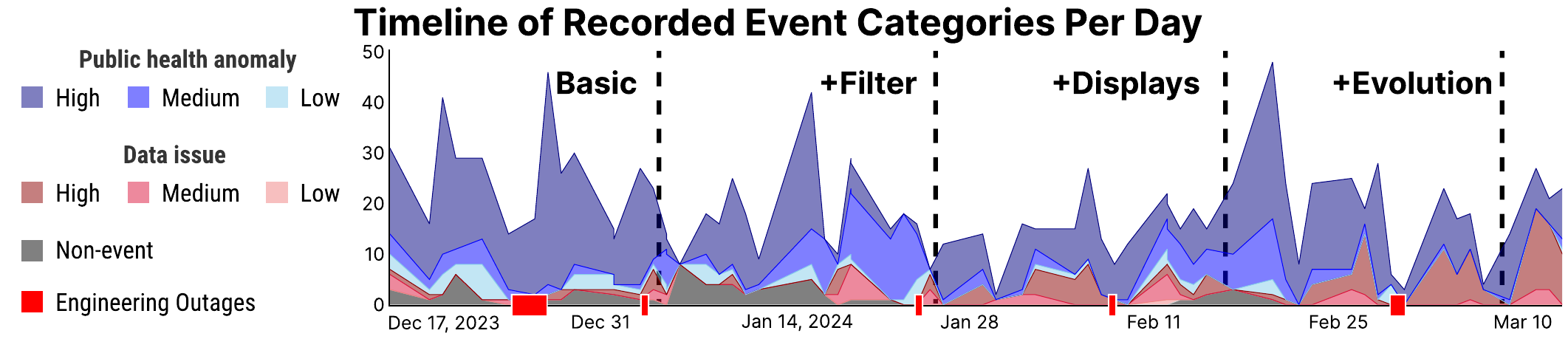}
 \caption{Recorded events per session (up to 49) are far greater than the 1-2 events reviewers would detect per session using {Baseline 1}. After filters were added, fewer rows were marked as a Non-event, suggesting that reviewers knew how to exclude data that was not interesting without complex synthesis strategies.}
 \label{fig:Timeline}
\end{figure}

In terms of output, the resulting documented events from the monitoring process, as shown in Fig. \ref{fig:Timeline}, were a mix of data quality and public health issues. Before filters were added to support the segmentation decision, there were many more points marked as a Non-event. However, after reviewers became more familiar with filters, they could exclude data that would generate high event scores but were not important or meaningful, like indicators that providers have stopped maintaining. Reviewers used filters on average $2.75 \pm 0.43$ times per day. Each filter can have up to 4 predicates (across signal, source, geo value, geo region), but reviewers only use an average of 1 predicate and only 1 value per predicate -- usually across geography or signal provider, once again supporting the desire for simple segmentation strategies instead of the fusion based norm in visualization. 

Because of the variance of the number of events detected in Fig. \ref{fig:Timeline}, online interfaces like the basic review interface are helpful because the number of rows reviewers will process (k) is \textit{unknown and unknowable} because reviewers are not required to use the system in any way -- their interest and engagement depend on their trust in the system and the number of important events that day.

\subsection{Lessons Learned and Guidelines:} 

The findings from this study underline the importance in designing and building a Human-centered AI system that matches the needs and constraints of the domain, as described here with public health monitoring stakeholders and an AI anomaly ranking method. Based on the relevance of our evaluation strategy  for practical adoption, we advocate more researchers in human-centered AI focus on longitudinal evaluation. For example, the data outages that sporadically occurred over months were unplanned, but are typical for data monitoring in practice and performance of the system would not have necessarily been tested on that setting. Because these results were measured under realistic conditions, Delphi's data reviewers have continued to use this system and approach, which demonstrates  promise for other practical applications. Our next steps aim to address  evolving stakeholder pain points from an interdisciplinary lens:
\begin{itemize}
\item Generalizing analysis across more data dimensions (e.g., age subgroups when available) \citep{rolka2007issues}
\item Improved meta-event detection by identifying event subsequences \citep{shmueli2010statistical}
\item Automating event summaries and generating narratives for stakeholders \citep{coletta2019can}
\end{itemize}
These next steps are well motivated given their importance to both Delphi's data reviewers and the field more generally \citep{burkom2017evolution}.

\section{Conclusion}
This work contributes an AI-based public health data monitoring system, with implications for large-scale data monitoring generally. This approach is designed to address the limitations of traditional alert-based methods at scale by using a collaborative design process involving engineers, computer scientist, and data reviewers with an AI ranking method. Our resulting user evaluation reveals that this approach boosts user engagement and detection rates and allows reviewers to correctly identify events \textbf{54x} faster than the previous deployed system (Baseline 1) and {6x} faster than the method alone (Baseline 2), as is relevant to the data reviewer's goal while monitoring public health data. 

\bibliographystyle{named}
\bibliography{ijcai25}

\end{document}